\documentclass{article}

\usepackage{amsfonts}
\usepackage{graphicx}        
\usepackage{multicol}        
\usepackage[bottom]{footmisc}
\usepackage{subfigure}
\usepackage[square,sort&compress,numbers]{natbib}

\usepackage{times,makeidx} 

\def\Aixi{{\sc Aixi} }
\def\Aixin{{\sc Aixi}}

\def\GM{G\"{o}del Machine }

\def\gmn{G\"{o}del machine}

\begin{document}
\title{2006: Celebrating 75 years of AI - \\ History and 
Outlook: the Next 25 Years\thanks{Invited contribution to the Proceedings of 
the ``50th Anniversary Summit of Artificial Intelligence'' at 
Monte Verita, Ascona, Switzerland, 9-14 July 2006 (variant accepted for
Springer's LNAI series)}}
\date{}
\author{J\"{u}rgen Schmidhuber \\
TU Munich, Boltzmannstr. 3,  85748 Garching bei M\"{u}nchen, Germany \& \\
IDSIA, Galleria 2, 6928 Manno (Lugano), Switzerland \\
{\tt juergen@idsia.ch - http://www.idsia.ch/\~{ }juergen}}

\maketitle

\begin{abstract}

When Kurt G\"{o}del layed the foundations of theoretical computer
science in 1931, he also introduced essential concepts of
the theory of Artificial Intelligence (AI).
Although much of subsequent AI research 
has focused on heuristics, 
which still play a major role in many 
practical AI applications, in the new millennium AI theory
has finally become a full-fledged formal science, with important 
optimality results for embodied agents living in unknown 
environments, obtained through a combination of
theory {\em \`{a} la} G\"{o}del 
and probability theory. 
Here we look back at important milestones of AI history, 
mention essential recent results,
and speculate about what we
may expect from the next 25 years, emphasizing the significance of
the ongoing dramatic hardware speedups, and discussing G\"{o}del-inspired,
self-referential, self-improving universal problem solvers.

\end{abstract}


\section{Highlights of AI History---From G\"{o}del to 2006}
\label{history}

{\bf G\"{o}del and Lilienfeld.} In 1931, 75 years ago and 
just a few years after Julius Lilienfeld patented the transistor,
Kurt G\"{o}del layed the foundations of theoretical computer
science (CS) with his work on universal formal languages and the limits
of proof and computation \cite{Goedel:31}. 
He constructed formal systems allowing
for self-referential statements that talk about themselves, in
particular, about whether they can be derived from a set of given
axioms through a computational theorem proving procedure.  G\"{o}del
went on to construct statements that claim their own unprovability,
to demonstrate that traditional math is either flawed in a certain
algorithmic sense or contains unprovable but true statements.

G\"{o}del's incompleteness result is widely regarded as the most
remarkable achievement of 20th century mathematics, although some
mathematicians say it is logic, not math, and others call it the
fundamental result of theoretical computer science, a discipline
that did not yet officially exist back then but was effectively
created through G\"{o}del's work.  It had enormous impact
not only on computer science but also on philosophy and other fields.
In particular, since humans can ``see'' the truth of G\"{o}del's unprovable
statements, some researchers mistakenly thought that his results
show that machines and Artificial Intelligences (AIs) will always 
be inferior to humans. Given the tremendous impact of
G\"{o}del's results on AI theory, it does make sense to date 
AI's beginnings back to his 1931 publication 75 years ago.

{\bf Zuse and Turing.}
In 1936 Alan Turing \cite{Turing:36} introduced the {\em Turing machine}
to reformulate G\"{o}del's results and Alonzo Church's extensions thereof.
TMs are often more convenient than G\"{o}del's integer-based formal systems,
and later became a central tool of CS theory. Simultaneously 
Konrad Zuse built the
first working program-controlled computers (1935-1941), using the
binary arithmetic and the {\em bits} of Gott\-fried Wilhelm von Leibniz (1701)
instead of the more cumbersome decimal system used by Charles Babbage,
who pioneered the concept of program-controlled computers 
in the 1840s, and tried to build one, although without success. 
By 1941, all the main ingredients of `modern' computer science were 
in place, a decade after G\"{o}del's paper,
a century after Babbage,
and roughly three centuries after Wilhelm Schickard, who started
the history of automatic computing hardware by constructing the
first non-program-controlled computer in 1623.

In the 1940s Zuse went on to devise the first high-level programming
language (Plankalk\"{u}l), which he used to write the first chess program.
Back then chess-playing was considered an intelligent activity,
hence one might call this chess program the first design of an AI program,
although Zuse did not really implement it back then. 
Soon afterwards, in 1948, Claude
Shannon \cite{Shannon:48} published information theory, recycling 
several older ideas such as Ludwig Boltzmann's entropy from 19th century 
statistical mechanics, and the {\em bit of information} (Leibniz, 1701). 

{\bf Relays, Tubes, Transistors.}
Alternative instances of transistors, the concept 
pioneered and patented by Julius Edgar Lilienfeld (1920s) 
and Oskar Heil (1935), were built by
William Shockley, Walter H. Brattain \& John Bardeen (1948: point contact transistor) 
as well as Herbert F. Matar\'{e} \& Heinrich Walker (1948,
exploiting transconductance effects of germanium diodes 
observed in the {\em Luftwaffe} during WW-II).
Today most transistors are of the field-effect type 
{\em \`{a} la} Lilienfeld \& Heil.
In principle a switch remains a switch no 
matter whether it is implemented
as a relay or a tube or a transistor, but
transistors switch faster than
relays (Zuse, 1941) and tubes (Colossus, 1943; ENIAC, 1946).
This eventually led to significant speedups of computer hardware, 
which was essential for many subsequent AI applications.

{\bf The I in AI.}
In 1950, some 56 years ago, Turing invented a famous subjective 
test to decide whether 
a machine or something else is intelligent. 
6 years later, and 25 years after G\"{o}del's paper, 
John McCarthy finally coined the term ``AI". 50 years later, in 2006, this
prompted some to celebrate the 50th birthday of AI, 
but this chapter's title should make clear that
its author cannot agree with this view---it is the 
thing that counts, not its name.

{\bf Roots of Probability-Based AI.}
In the 1960s and 1970s Ray Solomonoff 
combined theoretical CS and probability theory to
establish a general theory of universal inductive inference and predictive AI 
\cite{Solomonoff:78}
closely related to the concept of Kolmogorov
complexity \cite{Kolmogorov:65}.
His theoretically optimal predictors and their Bayesian learning
algorithms only assume that the observable reactions of the environment in
response to certain action sequences are sampled from
an unknown probability distribution contained in a set $M$ of all
enumerable distributions.  That is, given an observation sequence
we only assume there exists a computer program that can compute the
probabilities of the next possible observations.  This includes
all scientific theories of physics, of course. Since we typically
do not know this program, we predict using a weighted sum $\xi$ 
of {\em all} distributions in $\cal M$, where the sum of the weights does
not exceed 1.  It turns out that this is indeed the best one can
possibly do, in a very general sense \cite{Solomonoff:78,Hutter:04book+}.
Although the universal approach is practically infeasible since $M$
contains infinitely many distributions, it does represent the first 
sound and general theory of optimal prediction based on experience,
identifying the limits of both human and artificial predictors, and
providing a yardstick for all prediction machines to come.

{\bf AI vs Astrology?}
Unfortunately, failed prophecies of human-level AI with just a tiny fraction
of the brain's computing power discredited some of the AI research
in the 1960s and 70s. 
Many theoretical computer scientists actually regarded
much of the field with contempt for its perceived lack of hard theoretical results.
ETH Zurich's Turing award winner and creator of the PASCAL programming language,
Niklaus Wirth, did not hesitate to link AI to astrology. 
Practical AI of that era was dominated by rule-based expert systems and
Logic Programming. That is, despite Solomonoff's fundamental results,
a main focus of that time was on logical,
deterministic deduction of 
facts from previously known facts, as opposed to (probabilistic)
induction of hypotheses from experience. 

{\bf Evolution, Neurons, Ants.}
Largely unnoticed by mainstream AI gurus of that era,
a biology-inspired type of AI emerged in the 1960s when
Ingo Rechenberg pioneered the method of artificial evolution 
to solve complex optimization tasks \cite{Rechenberg:71}, 
such as the design of optimal airplane wings
or combustion chambers of rocket nozzles.
Such methods (and later variants thereof, e.g., Holland \cite{Holland:75} (1970s),
often gave better results than classical approaches.
In the following decades, other types of ``subsymbolic" AI also 
became popular, especially neural networks. Early neural net papers include 
those of McCulloch \& Pitts, 1940s (linking certain simple
neural nets to old and well-known, simple mathematical concepts such
as linear regression); 
Minsky \& Papert \cite{MinskyPapert:69} (temporarily discouraging
neural network research), 
Kohonen \cite{Kohonen:88}, Amari, 1960s;
Werbos \cite{Werbos:74}, 1970s; and many others in the 1980s. 
Orthogonal approaches included fuzzy logic (Zadeh, 1960s), 
Rissanen's practical variants \cite{Rissanen:78}
of Solomonoff's universal method,
``representation-free" AI (Brooks \cite{broo91a}), 
Artificial Ants (Dorigo \& Gambardella \cite{Dorigo:99}, 1990s), 
statistical learning theory (in less general settings than those
studied by Solomonoff) \& support vector machines 
(Vapnik \cite{Vapnik:95} and others).
As of 2006, this alternative type of AI research
is receiving more attention than
``Good Old-Fashioned AI'' (GOFAI).

{\bf Mainstream AI Marries Statistics.}
A dominant theme of the 1980s and 90s was the marriage of 
mainstream AI and old concepts from probability theory.
Bayes networks, Hidden Markov Models, and numerous other probabilistic
models found wide applications ranging from pattern recognition,
medical diagnosis, data mining, machine translation, robotics, etc.

{\bf Hardware Outshining Software: Humanoids, Robot Cars, Etc.}
In the 1990s and 2000s, much of the progress in practical AI was
due to better hardware, getting roughly 1000 times faster per Euro
per decade.  In 1995, a fast vision-based robot car by Ernst Dickmanns
(whose team built the world's first reliable robot cars in the early 1980s
with the help of Mercedes-Benz, e. g., \cite{Dickmanns:94})
autonomously drove 1000 miles from Munich to Denmark
and back, in traffic at up to 120 mph, automatically passing other 
cars (a safety driver took over only rarely in critical situations). 
Japanese labs (Honda, Sony) and Pfeiffer's lab at TU Munich 
built famous humanoid walking robots. Engineering problems often seemed 
more challenging than AI-related problems. 

Another source of progress was the dramatically improved
access to all kinds of data through the WWW,
created by Tim Berners-Lee at the European particle
collider CERN (Switzerland) in 1990. This greatly facilitated 
and encouraged all kinds of ``intelligent'' data mining applications.
However, there were few if any obvious
fundamental algorithmic breakthroughs; improvements /
extensions of already existing algorithms seemed less impressive
and less crucial than hardware advances. For example, chess world
champion Kasparov was beaten by a fast IBM computer running a fairly
standard algorithm. Rather simple but computationally expensive
probabilistic methods for speech recognition, statistical machine
translation, computer vision, optimization, virtual realities etc. 
started to become
feasible on PCs, mainly because PCs had become 1000 times more powerful
within a decade or so.

{\bf 2006.}
As noted by Stefan Artmann (personal communication, 2006), today's AI textbooks
seem substantially more complex and less unified than those of several 
decades ago, e. g.,  \cite{Nilsson:80},
since they have to cover so many apparently quite different subjects.
There seems to be a need for a new unifying view of intelligence.
In the author's opinion this view already exists, as will be discussed below.

\section{Subjective Selected Highlights of Present AI}
\label{present}

The more recent some event, the harder it is to judge its 
long-term significance. But this biased author thinks that 
the most important thing that happened recently in AI is 
the begin of a
transition from a heuristics-dominated science 
(e.g., \cite{SOAR:93})
to a real formal science. Let us elaborate on this topic.

\subsection{The Two Ways of Making a Dent in AI Research}
\label{dent}

There are at least two convincing ways of doing AI research:
{\bf (1)} 
construct a (possibly heuristic) machine or algorithm that 
somehow (it does not really matter how)
solves a previously unsolved interesting problem, such as
beating the best human player of {\em Go} (success
will outshine any lack of theory). 
Or {\bf (2)} 
prove that a particular novel algorithm is optimal for 
an important class of AI problems.

It is the nature of heuristics (case {\bf (1)}) that they 
lack staying power, as they may soon get replaced by next 
year's even better heuristics.  
Theorems (case {\bf (2)}), however, are for eternity. 
That's why formal sciences prefer theorems.

For example, probability theory became a formal
science centuries ago, and totally formal in 1933 with Kolmogorov's
axioms \cite{Kolmogorov:33}, shortly after 
G\"{o}del's paper \cite{Goedel:31}. 
Old but provably optimal
techniques of probability theory are still in every day's use,
and in fact highly significant for modern AI, while many
initially successful heuristic approaches eventually became 
unfashionable, of interest mainly to the historians of the field.

\subsection{
No Brain Without a Body /
AI Becoming a Formal Science 
}
\label{formal}

Heuristic approaches will continue to play an
important role in many AI applications, to the extent they
empirically outperform competing methods. But 
like with all young sciences at the transition point
between an early intuition-dominated and a later formal era,
the importance of mathematical optimality theorems is
growing quickly. Progress in the formal era,
however, is and will be driven by a different breed of researchers, 
a fact that is not necessarily universally 
enjoyed and welcomed by all the earlier pioneers.

Today the importance of embodied, embedded AI is almost
universally acknowledged (e. g., \cite{Pfeifer:01}),
as obvious from frequently
overheard remarks such as ``let the physics compute'' and ``no
brain without a body.'' Many present AI researchers
focus on real robots living in real physical environments. 
To some of them
the title of this subsection may seem oxymoronic:
the extension of AI into the realm  of the physical  body seems
to be a step away from  formalism.
But the new millennium's formal point of view 
is actually taking this step into account in a very general way,
through the first mathematical theory of 
universal embedded AI, combining ``old" theoretical 
computer science and ``ancient"
probability theory to derive optimal behavior for embedded,
embodied  rational agents living in unknown but learnable
environments.  More on this below.

\subsection{What's the I in AI?  What is Life? Etc.}
\label{what}

Before we proceed, let us clarify what we are talking about.
Shouldn't researchers on 
Artificial Intelligence (AI) and Artificial Life (AL)
agree on basic questions such as: What is Intelligence? What is Life? 
Interestingly they don't.

{\bf Are Cars Alive?} 
For example, AL researchers often offer definitions of life such as:
it must reproduce, evolve, etc.  Cars are alive, too, according to most of these 
definitions.  For example, cars evolve and multiply. 
They need complex environments with car factories to do so, but living animals 
also need complex environments full of chemicals and other animals 
to reproduce --- the DNA information by itself does not suffice.
There is no obvious fundamental difference between an organism 
whose self-replication information is stored in its DNA, and a car
whose self-replication information is stored in a car builder's manual
in the glove compartment. To copy itself, the organism needs its mothers 
womb plus numerous other objects and living beings in its 
environment (such as trillions of bacteria inside and outside of
the mother's body).  The car needs iron mines and car part 
factories and human workers. 

{\bf What is Intelligence?} If we cannot agree on what's life, or, for that matter, love, 
or consciousness (another fashionable topic),
how can there be any hope to define intelligence? Turing's definition (1950, 19 
years after G\"{o}del's paper) was totally subjective: intelligent is what
convinces me that it is intelligent while I am interacting with it. 
Fortunately, however,
there are more formal and less subjective definitions.

\subsection{Formal AI Definitions}
\label{formaldef}

Popper said: all life is problem solving \cite{Popper:99}.
Instead of defining intelligence in Turing's rather vague and subjective
way we define intelligence with respect to the abilities of universal
optimal problem solvers.

Consider a learning robotic agent with a single life which
consists of discrete cycles or time steps $t=1, 2, \ldots, T$.
Its total lifetime $T$ may or may not be known in advance.
In what follows,the value of any time-varying variable $Q$
at time $t$ ($1 \leq t \leq T$) will be denoted by $Q(t)$,
the ordered sequence of values $Q(1),\ldots,Q(t)$ by $Q(\leq t)$,
and the (possibly empty) sequence $Q(1),\ldots,Q(t-1)$ by $Q(< t)$.

At any given $t$  the robot receives a real-valued input vector $x(t)$ from
the environment and executes a real-valued 
action $y(t)$ which may affect future inputs; at times $t<T$ its goal
is to maximize future success or {\em utility}
\begin{equation}
\label{u}
u(t) =
E_{\mu} \left [ \sum_{\tau=t+1}^T  
r(\tau)~~ \Bigg| ~~ h(\leq t) \right ],
\end{equation}
where $r(t)$ is an additional real-valued reward input at time $t$,
$h(t)$ the ordered triple $[x(t), y(t), r(t)]$
(hence $h(\leq t)$ is the known history up to $t$),
and $E_{\mu}(\cdot \mid \cdot)$ denotes the conditional expectation operator
with respect to some possibly unknown distribution $\mu$ from a set $M$
of possible distributions. Here $M$ reflects
whatever is known about the possibly probabilistic reactions
of the environment.  For example, $M$ may contain all computable
distributions \cite{Solomonoff:78,Hutter:04book+}.
Note that unlike in most previous work by others \cite{Sutton:98},
there is just one life, no need for predefined repeatable trials, 
no restriction to Markovian 
interfaces between sensors and environment,
and the utility function implicitly takes into account the 
expected remaining lifespan $E_{\mu}(T \mid  h(\leq t))$
and thus the possibility to extend it through appropriate actions
\cite{Schmidhuber:05gmai}.

Any formal problem or sequence of problems can be encoded in
the reward function. For example, the reward functions of many
living or robotic beings cause occasional hunger or pain or pleasure
signals etc.  At time $t$ an optimal AI will make the best possible
use of experience $h(\leq t)$ to maximize $u(t)$. But how?

\subsection{Universal, Mathematically Optimal, But Incomputable AI}
\label{unilearn}

Unbeknownst to many traditional AI researchers,
there is indeed an extremely general ``best'' way of 
exploiting previous experience.
At any time $t$, the recent theoretically optimal
yet practically infeasible
reinforcement learning (RL) algorithm \Aixi \cite{Hutter:04book+} 
uses Solomonoff's above-mentioned universal prediction scheme  
to select those action sequences that promise maximal
future reward up to some horizon,
given the current data $h(\leq t)$.  
Using a variant of Solomonoff's universal
probability mixture $\xi$, in cycle $t+1$, \Aixi
selects as its next action the first action of an action sequence
maximizing $\xi$-predicted reward up to the horizon. 
Hutter's recent work \cite{Hutter:04book+} demonstrated \Aixin's optimal
use of observations as follows.  The Bayes-optimal policy $p^\xi$ based on
the mixture $\xi$ 
is self-optimizing in the sense that its average
utility value converges asymptotically for all $\mu \in \cal M$ to the
optimal value achieved by the (infeasible) Bayes-optimal policy $p^\mu$
which knows $\mu$ in advance.  The necessary condition that $\cal M$
admits self-optimizing policies is also sufficient.

Of course one cannot claim the old AI is devoid of formal research! 
The recent approach above, however, goes far beyond previous formally justified  
but very limited AI-related approaches ranging from linear perceptrons 
\cite{MinskyPapert:69} to the $A^*$-algorithm \cite{Nilsson:80}.
It provides, for the first time, a mathematically sound theory 
of general AI and optimal decision making based on experience, identifying the limits of
both human and artificial intelligence, and a yardstick
for any future, scaled-down, practically feasible approach to general AI.

\subsection{Optimal Curiosity and Creativity}
\label{curious}

No theory of AI will be convincing if it does not 
explain curiosity and creativity, which many consider
as important ingredients of intelligence.
We can provide an explanation in the framework of
optimal reward maximizers such as those from the previous subsection.

It is possible to come up with theoretically optimal ways of 
improving the predictive world model of
a curious robotic agent \cite{Schmidhuber:06cs}, 
extending earlier ideas on how to implement artificial curiosity
\cite{Schmidhuber:91singaporecur}:
{\em The rewards of an optimal reinforcement learner are the predictor's
improvements} on the observation history so far.
They encourage the 
reinforcement learner to produce action sequences
that cause the creation and the learning of new, 
previously unknown regularities in the sensory input stream.
It turns out that art and creativity can be explained as by-products of such
intrinsic curiosity rewards:  good observer-dependent 
art deepens the observer's insights about this world or
possible worlds, connecting previously disconnected patterns 
in an initially surprising way that eventually becomes known and boring. 
While previous attempts at describing what is satisfactory 
art or music were informal,  this work
permits the first {\em technical, formal}
approach to understanding the nature of 
art and creativity \cite{Schmidhuber:06cs}.

\subsection{Computable, Asymptotically Optimal General Problem Solver}
\label{fast}

Using the Speed Prior \cite{Schmidhuber:02colt}
one can scale down the universal approach above such 
that it becomes computable.  In what follows we will mention 
general methods whose optimality criteria explicitly take into account 
the computational costs of prediction and decision making---compare
\cite{Levin:73}.

The recent asymptotically
optimal search algorithm for {\em all} well-defined problems
\cite{Hutter:04book+}  allocates part of the total
search time to searching the space of proofs for provably correct
candidate programs with provable upper runtime bounds;
at any given time it
focuses resources on those programs with the currently
best proven time bounds.
The method is as fast as the initially
unknown fastest problem solver for the given problem class,
save for a constant slowdown factor of at most
$1 + \epsilon$, $\epsilon >0$, and an additive
constant that does not depend on the problem instance!

Is this algorithm then the {\em holy grail} of computer science?
Unfortunately not quite, since the additive constant (which disappears
in the $O()$-notation of theoretical CS) may be huge,
and practical applications may not ignore it.
This motivates the next section, which
addresses all kinds of formal optimality
(not just asymptotic optimality).

\subsection{Fully Self-Referential, Self-Improving \GM}
\label{gm}

We may use G\"{o}del's self-reference trick to build a universal
general, fully self-referential, 
self-improving, optimally efficient problem solver
\cite{Schmidhuber:05gmai}.
A \GM is a computer whose original
software includes axioms describing the hardware and the original
software (this is possible without circularity) plus whatever is known
about the (probabilistic) environment plus some formal
goal in form of an arbitrary user-defined utility function, e.g.,
cumulative future expected reward in a sequence of optimization
tasks - see equation (\ref{u}). The original software also includes a 
proof searcher which uses the axioms 
(and possibly an online variant of Levin's universal 
search \cite{Levin:73})
to systematically make pairs (``proof'', ``program'')
until it finds a proof that a rewrite of the original software
through ``program'' will increase utility. The machine can be designed
such that each self-rewrite is necessarily globally optimal in the
sense of the utility function, even those rewrites that destroy the
proof searcher
\cite{Schmidhuber:05gmai}.

\subsection{Practical Algorithms for Program Learning}
\label{programs}

The theoretically optimal universal methods above 
are optimal in ways that
do not (yet) immediately yield practically feasible
general problem solvers, due to possibly large initial
overhead costs.
Which are today's practically most promising 
extensions of traditional machine learning?

Since virtually all realistic sensory inputs of robots and other
cognitive systems are sequential by nature, the future of machine 
learning and AI in general depends on progress in
in sequence processing as opposed to the traditional processing
of stationary input patterns.
To narrow the gap between learning abilities of humans and
machines, we will have to study how
to learn general algorithms instead of such reactive mappings.
Most traditional methods for learning time series and mappings from
sequences to sequences, however, are based on simple time windows:
one of the numerous feedforward ML techniques such as
feedforward neural nets (NN) \cite{Bishop:95}
or support vector machines \cite{Vapnik:95} is used to map
a restricted, fixed time window of sequential input values to
desired target values.  Of
course such approaches are bound to fail if
there are temporal dependencies exceeding the time window size.
Large time windows, on the other hand, yield
unacceptable numbers of free parameters.

Presently studied, rather general sequence learners include
certain probabilistic approaches and especially 
recurrent neural networks (RNNs), e.g.,
\cite{Pearlmutter:95}.
RNNs have adaptive feedback
connections that allow them to learn mappings from input sequences
to output sequences. They can implement any
sequential, algorithmic behavior implementable on a personal computer. 
In gradient-based RNNs, however, we can {\em differentiate our wishes with
respect to programs,} to obtain a search direction
in algorithm space. RNNs
are biologically more plausible and computationally more
powerful than other adaptive models such as
Hidden Markov Models (HMMs - no continuous internal states),
feedforward networks \& Support Vector Machines (no internal states at all).
For several reasons, however,
the first RNNs could not learn to look far back into the past. 
This problem was overcome by RNNs of the {\em Long Short-Term Memory} type (LSTM),
currently the most powerful and practical 
supervised RNN architecture for many applications, trainable either by gradient descent  
\cite{Hochreiter:97lstm} 
or evolutionary methods \cite{Schmidhuber:06nc}, occasionally profiting
from a marriage with probabilistic approaches
\cite{Graves:06icml}.

Unsupervised RNNs that learn without a teacher 
to control physical processes or robots 
frequently use 
evolutionary algorithms \cite{Rechenberg:71,Holland:75} 
to learn appropriate programs (RNN weight matrices) through trial and
error \cite{yao:review93}.
Recent work brought progress through a focus on reducing
search spaces by co-evolving the comparatively small
weight vectors of individual recurrent neurons
\cite{Gomez:03+}.
Such RNNs can learn to create
memories of important events, solving numerous
RL / optimization tasks unsolvable by traditional RL methods
\cite{Gomez:03+,Gomez:06ecml}. They are 
among the most promising methods for practical program learning, and
currently being applied to the control of sophisticated
robots such as the 
walking biped of TU Munich \cite{Lohmeier:04}.

\section{The Next 25 Years}
\label{next25}
Where will AI research stand in 2031, 25 years from now, 100 years after
G\"{o}del's ground-breaking paper \cite{Goedel:31}, some 200 years after 
Babbage's first designs, some 400 years after the first automatic 
calculator by Schickard 
(and some 2000 years after the crucifixion of the man whose 
birth year anchors the Western calendar)?

{\bf Trivial predictions} are those that just naively extrapolate
the current trends, such as: computers will continue to 
get faster by a factor of roughly 1000 per decade; 
hence they will be at least a million times faster 
by 2031. According to frequent estimates, current supercomputers 
achieve roughly 1 percent of the raw computational power of
a human brain, hence those of 2031 will have 10,000 ``brain
powers''; and even cheap devices will achieve many
brain powers.  Many tasks that are hard for today's 
software on present machines will become easy 
without even fundamentally changing the algorithms.
This includes numerous pattern recognition and
control tasks arising in factories of many industries,
currently still employing humans instead of robots.

Will theoretical advances and practical software keep up with the
hardware development?  We are convinced they will. As discussed
above, the new millennium has already brought fundamental new 
insights into the problem of constructing theoretically optimal 
rational agents or universal AIs, even if those do not yet immediately
translate into practically feasible methods.  On
the other hand, on a more practical level, there has been rapid
progress in learning algorithms for agents interacting with a dynamic
environment, autonomously discovering true sequence-processing,
problem-solving programs, as opposed to the reactive mappings from
stationary inputs to outputs studied in most of traditional machine
learning research.  In the author's opinion the above-mentioned
theoretical and practical strands are going to converge. In conjunction
with the ongoing hardware advances this will yield non-universal but 
nevertheless rather general
artificial problem-solvers whose capabilities will exceed those of 
most if not all humans in many domains of commercial interest. 
This may seem like a bold prediction to some, but it is actually a
trivial one as there are so many experts who would agree with it.

{\bf Nontrivial predictions} are those that anticipate
truly unexpected, revolutionary breakthroughs.  By definition, these are 
hard to predict.
For example, in 1985 only very few scientists and
science fiction authors predicted the
WWW revolution of the 1990s. The few who did
were not influential enough to make a significant
part of humanity 
believe in their predictions and prepare for their coming true.
Similarly, after the latest stock market crash one
can always find with high probability some ``prophet in the desert'' who 
predicted it in advance, but had few if any followers
until the crash really occurred.

Truly nontrivial predictions are those 
that most will not believe until they come true.
We will mostly restrict ourselves
to trivial predictions like those above and
refrain from too much speculation in form of nontrivial 
ones.  However, we may have a look at previous
unexpected scientific breakthroughs and try to discern 
a pattern, a pattern that may not allow us to precisely
predict the details of the next revolution
but at least its timing.

\subsection{A Pattern in the History of Revolutions?}

Let us put the AI-oriented developments \cite{Schmidhuber:06ai}
discussed above in a broader context, and
look at the history of major scientific revolutions
and essential historic developments (that is, 
the subjects of the major chapters in history books) 
since the beginnings of modern man over 40,000 years ago 
\cite{Schmidhuber:06newmillenniumai,Schmidhuber:06history}.
Amazingly, they seem to match 
a binary logarithmic scale marking exponentially declining temporal intervals
\cite{Schmidhuber:06newmillenniumai},
each half the size of the previous one, and measurable 
in terms of powers of 2 multiplied by a human lifetime 
(roughly 80 years---throughout recorded history many individuals
have reached this age, although the average lifetime often
was shorter, mostly  due to high children mortality).
It looks as if history itself will {\em converge} in a 
historic singularity or Omega point $\Omega$ around $2040$ 
(the term {\em historic singularity} is apparently
due to Stanislaw Ulam (1950s) and 
was popularized by Vernor Vinge \cite{Vinge:93} in the 1990s). 
To convince yourself of history's convergence, 
associate an error bar of not much 
more than 10 percent with each date below:
\begin{enumerate}
\item
$\Omega - 2^9$ lifetimes: 
modern humans start colonizing
the world from Africa 
\item
$\Omega - 2^8$ lifetimes: 
bow and arrow invented; hunting revolution
\item
$\Omega - 2^7$ lifetimes: 
invention of agriculture; first
permanent settlements; beginnings of civilization
\item
$\Omega - 2^6$ lifetimes: 
first high civilizations (Sumeria, Egypt),
and the most important invention of recorded history,
namely, the one that made recorded history possible: writing 
\item
$\Omega - 2^5$ lifetimes: 
the ancient Greeks invent democracy and
lay the foundations of Western science and art and
philosophy, from algorithmic procedures and formal proofs to
anatomically perfect sculptures, harmonic music, and organized sports.
Old Testament written (basis of Judaism, Christianity, Islam); 
major Asian religions founded.
High civilizations in China, origin of the first 
calculation tools, and India, origin of alphabets and the zero
\item
$\Omega - 2^4$ lifetimes: 
bookprint (often called the most important invention of
the past 2000 years) invented in China. 
Islamic science and culture start spreading across 
large parts of the known world (this has sometimes been 
called the most important event between Antiquity and
the age of discoveries)
\item
$\Omega - 2^3$ lifetimes: 
the Mongolian Empire, the largest and most dominant 
empire ever (possibly including most of
humanity and the world economy),
stretches across Asia from Korea all the way to Germany.
Chinese fleets and later also European vessels start
exploring the world. Gun powder and guns invented in China.
Rennaissance and Western bookprint 
(often called the most influential
invention of the past 1000 years)
and subsequent Reformation 
in Europe. Begin of the Scientific Revolution 
\item
$\Omega - 2^2$ lifetimes: 
Age of enlightenment and rational thought in Europe.
Massive progress in the sciences; first flying machines;
first steam engines prepare the industrial revolution 
\item
$\Omega - 2$ lifetimes: 
Second industrial revolution based on
combustion engines, cheap electricity, and modern chemistry.
Birth of modern medicine through the
germ theory of disease;
genetic and evolution theory.
European colonialism at its short-lived peak
\item
$\Omega - 1$ lifetime: 
modern post-World War II society and pop culture emerges; 
superpower stalemate based on nuclear deterrence.
The 20th century super-exponential population explosion (from
1.6 billion to 6 billion people, mainly
due to the Haber-Bosch process \cite{Smil:99}) is at its peak.
First spacecraft and commercial computers;
DNA structure unveiled 
\item
$\Omega - 1/2$ lifetime (now): 
for the first time in history most of the most
destructive weapons are dismantled, after the Cold War's peaceful end.
3rd industrial revolution based on
personal computers and the World Wide Web.
A mathematical theory of universal AI emerges (see sections above) - 
will this be considered a milestone in the future?
\item
$\Omega - 1/4$ lifetime: 
This point will be reached around 2020. By then many computers
will have substantially more raw computing power than human brains.
\item
$\Omega - 1/8$ lifetime (100 years after G\"{o}del's paper):  will
practical variants of \gmn s start a runaway evolution of continually
self-improving superminds way beyond human imagination,
causing far more unpredictable
revolutions in the final decade before $\Omega$ 
than during all the millennia before?
\item
...
\end{enumerate}

The following disclosure should help the reader to take this list
with a grain of salt though. The author, who admits being very
interested in witnessing $\Omega$, was born in 1963, and therefore
perhaps should not expect to live long past 2040.  This may motivate
him to uncover certain historic patterns that fit his desires, while
ignoring other patterns that do not.  Perhaps there even is a general
rule for both the individual memory of single humans and the
collective memory of entire societies and their history books:
constant amounts of memory space get allocated to exponentially
larger, adjacent time intervals further and further into the past.
Maybe that's why there has never been a shortage of prophets
predicting that the end is near - the important events according
to one's own view of the past always seem to accelerate exponentially.
See \cite{Schmidhuber:06newmillenniumai} for
a more thorough discussion of this possibility.

\bibliography{bib}
\bibliographystyle{plain}
\end{document}